\documentclass[conference]{IEEEtran}
\IEEEoverridecommandlockouts
\usepackage{cite}
\usepackage{amsmath,amssymb,amsfonts}
\usepackage{algorithmic}
\usepackage{graphicx}
\usepackage{textcomp}
\usepackage{xcolor}
\usepackage{booktabs}
\usepackage{siunitx}
\usepackage{url}
\usepackage{subfigure}
\usepackage[font=small]{caption}

\def\BibTeX{{\rm B\kern-.05em{\sc i\kern-.025em b}\kern-.08em
    T\kern-.1667em\lower.7ex\hbox{E}\kern-.125emX}}
\begin{document}

\title{Implementation and Evaluation of Networked Model Predictive Control System on Universal Robot}

\author{\IEEEauthorblockN{1\textsuperscript{st} Mahsa Noroozi}
\IEEEauthorblockA{\textit{Institute of Communications Technology} \\
\textit{Leibniz Universität Hannover}\\
Hanover, Germany \\
mahsa.noroozi@ikt.uni-hannover.de}
\and
\IEEEauthorblockN{2\textsuperscript{nd} Kai Wang}
\IEEEauthorblockA{\textit{Institute of Communications Technology} \\
\textit{Leibniz Universität Hannover}\\
Hanover, Germany \\
kai.wang@stud.uni-hannover.de}
}

\maketitle
\thispagestyle{plain} 
\pagestyle{plain}

\begin{abstract}
Networked control systems are closed-loop feedback control systems containing system components that may be distributed geographically in different locations and interconnected via a communication network such as the Internet. The quality of network communication is a crucial factor that significantly affects the performance of remote control. This is due to the fact that network uncertainties can occur in the transmission of packets in the forward and backward channels of the system. The two most significant among these uncertainties are network time delay and packet loss. To overcome these challenges, the networked predictive control system has been proposed to provide improved performance and robustness using predictive controllers and compensation strategies. In particular, the model predictive control method is well-suited as an advanced approach compared to conventional methods. In this paper, a networked model predictive control system consisting of a model predictive control method and compensation strategies is implemented to control and stabilize a robot arm as a physical system. In particular, this work aims to analyze the performance of the system under the influence of network time delay and packet loss. Using appropriate performance and robustness metrics, an in-depth investigation of the impacts of these network uncertainties is performed. Furthermore, the forward and backward channels of the network are examined in detail in this study.
\end{abstract}


\section{Introduction}
\label{Section1}
\begin{figure}[htb]
\centering
  \includegraphics[width=1\linewidth]{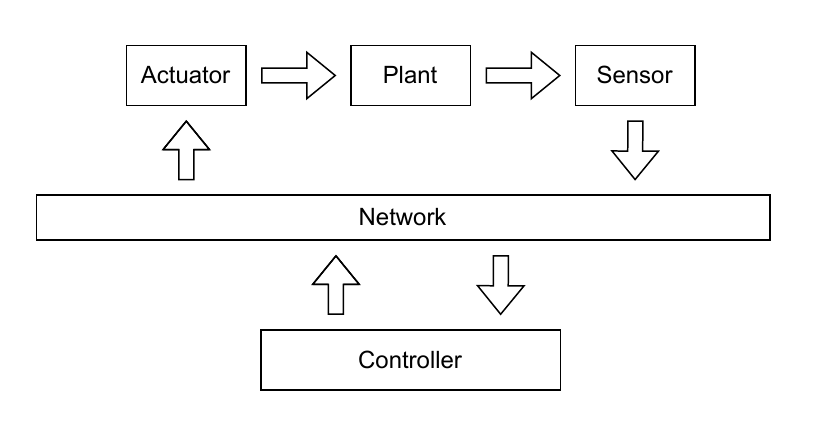}
  \caption{Basic architecture of an NCS}
  \label{NCS}
\end{figure}

Cyber-Physical System (CPS) describes the future networking of the physical world of machines, plants, and devices with the virtual world of the Internet or cyberspace. An example of such a system is the Networked Control System (NCS). NCS is a feedback closed-loop control system that connects system components such as actuators, plants, sensors, and controllers via a communication network like the Internet. Fig~\ref{NCS} shows the architecture of an NCS. Process control, vehicle industry, industrial automation, and robot manipulators are different areas in NCSs \cite{li2019challenging,zhang2019networked}. Using a communication network in an NCS architecture has both advantages and disadvantages. Low installation and maintenance costs, increased system flexibility, and reduced wiring requirements are some advantages of NCS. However, due to the limited capacity of the communication network, it also has some weaknesses resulting from network congestion. Some of these limitations are network time delay and packet loss, which can affect performance and cause instability in NCSs.

There are various studies that deal with the effects of network imperfections to stabilize the systems. They are categorized as control of the network and control over the network. The former is concerned with the study of communication networks to make them suitable for real-time NCS. In the latter, various control methods such as the optimal stochastic control method \cite{nilsson1998real}, the event-based method \cite{xi1998planning}, and the predictive control method \cite{liu2004networked} are used to mitigate the adverse effects of the network parameters. \cite{zhang2012design} investigated the design and implementation of Networked Predictive Control (NPC) for the first time. NPC systems use predictive controllers as an active compensation method in NCSs. Model Predictive Control (MPC) is a well-known predictive controller in this field \cite{kouvaritakis2016model}.

Round Trip Time (RTT) delay is used to measure the delays in end-to-end communication. It is the total time delay of a control cycle. Considering RTT eliminates the need for clock synchronization between two sides, i.e., the plant side and the controller side. \cite{li2018robust,vafamand2018networked,pang2022active} based their prediction and delay compensation on the RTT delay. \cite{chen2011stability} used network predictive control to teleoperate a robot with variable RTT delay using a sparse multi-variable linear regression algorithm to predict the RTT delay. Despite the simplicity of using RTT delay, forward and backward channels are distinguished from each other in NCSs. The controller doesn't know the time that the control signal needs to reach the plant in the forward channel. Therefore, an exact correction is not possible while calculating the control signals.

Considering a universal robot as a plant, \cite{omarali2017real} developed a real-time teleoperation system through human motion capture with a visualization utility. A linear explicit model predictive robot control is implemented for the online generation of optimal robot trajectories that match the operator’s wrist position and orientation without considering the network constraints. \cite{ali2017mpc} performed a comparison of PID (Proportional Integral and Derivative) and MPC algorithms on an upper limb rehabilitation robot with three degrees of freedom. A new networked predictive control system was implemented in \cite{noroozi2022performance} to control a universal robot arm with network constraints such as network time delay and packet loss. In \cite{gold2022catching}, a nonlinear model predictive control-based planning and control approach for catching objects in flight with a robot arm was presented, where this model doesn't impose network constraints such as network time delay.

Due to real-world network constraints, we implemented a Networked Model Predictive Control (NMPC) system to control and stabilize a universal robot arm while the network experiences time delay and packet loss. In this system, a model predictive controller is implemented along with network compensation methods. We analyze the performance of our system under variations of the parameters and compare it with an NCS using a PID controller. In this work, we consider both cases, i.e., a separate forward channel and a backward channel, and both together as RTT.

The remainder of this paper is organized as follows. First, section~\ref{Section2} describes the analytical model of the system that we have implemented. In section~\ref{Section3}, the evaluation results with different parameters in different cases are shown and a comparison with another system is made. Finally, section~\ref{Section4} concludes this paper.

\section{Analytical model}
\label{Section2}
\begin{figure}[htb]
\centering
  \includegraphics[width=1\linewidth]{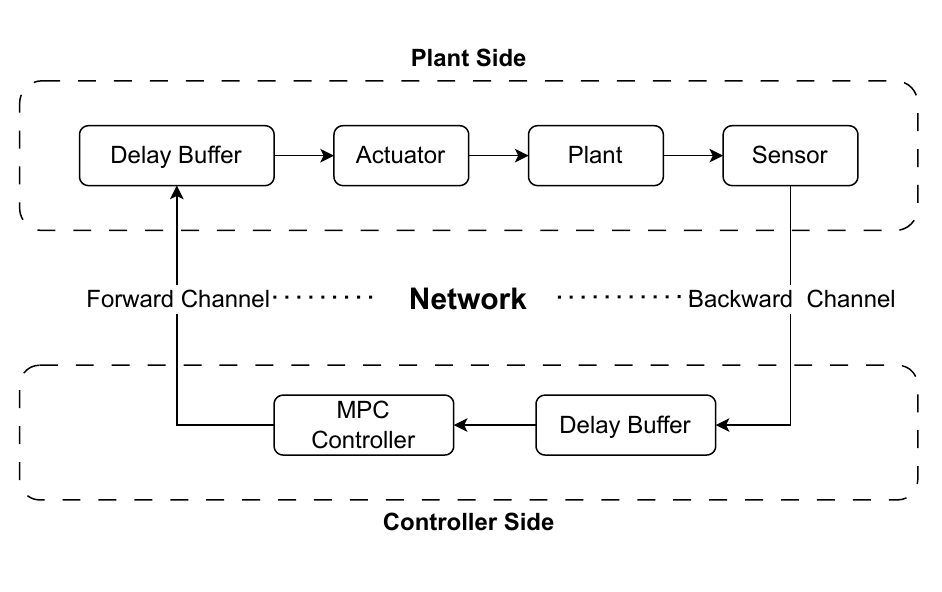}
  \caption{Basic architecture of the NMPC system}
  \label{NPC}
\end{figure}

In this work, an NMPC system is developed using an MPC approach. Compared to the basic structure of the NCS, it has some additional components, which are shown in Fig~\ref{NPC}. This model consists of two buffers on both sides as a compensator, an MPC as a controller, and separate channels for forward and backward channels. 

\subsection{Plant model}
The plant used in this work is a 6-axis universal robot UR5e. This robot arm is controlled by acceleration in three dimensions. For simplicity, we use one dimension since all dimensions behave the same. The inverse kinematics of the robot is handled locally and the position indicator of the universal robot is accurate to $0.03mm$ \cite{pollak2020measurement}. The system description of the plant as a linear system in state space description is: 
\begin{equation}
\centering
    \dot{\boldsymbol{x}}(t) = \boldsymbol{A}\boldsymbol{x}(t)+\boldsymbol{B}\boldsymbol{u}(t)
    = \begin{bmatrix}0 & 1\\0 & 0\end{bmatrix}\boldsymbol{x}(t)
    + \begin{bmatrix}0 \\ 1 \end{bmatrix}\boldsymbol{u}(t)
    \label{2}
\end{equation}
with $\boldsymbol{x}(t)$ as the state consisting of the angle ($\theta (t)$) and the velocity ($\dot{\theta} (t)$), and $\boldsymbol{u}(t)$ as the control input:
\begin{equation}
    \centering
    \boldsymbol{x}(t) = \begin{bmatrix}
        \theta (t)
        \\
        \dot{\theta} (t)
    \end{bmatrix}
\end{equation}
The output of the system is:
\begin{equation}
\centering
    \boldsymbol{y}(t) = \boldsymbol{C}\boldsymbol{x}(t) = \begin{bmatrix}1 & 0\\0 & 1\end{bmatrix}\boldsymbol{x}(t)
\end{equation}

\subsection{Model predictive control}
Given that the feedback states in our plant model are defined as angle and angular velocity, our optimization problem can be designed as follows:

\begin{equation*}
\begin{aligned}
\displaystyle{\min_{u}} J
&= \int (x(t)-r(t))^T Q_x (x(t)-r(t)) + u(t)^T Q_u u(t) dt \\
&= \sum_{i=1}^{N} (x(i)-r(i))^T Q_x (x(i)-r(i)) + u(i)^T Q_u u(i) \\
\end{aligned}
\end{equation*}
The vector $r$ represents the target state of the joint. The cost function defined here is composed of the part of the error between the current state and the desired state and the part of the input effort. Each of these parts is weighted separately by its respective weight matrix or weight value. Here, $Q_x$ is the matrix for the state error and $Q_u$ is the matrix value for the input effort. These are defined as follows:
\begin{equation*}
Q_x = \begin{bmatrix} 13.0\\1.8 \end{bmatrix};  \quad
Q_u = 0.01
\end{equation*}
which were determined by iterative tests. The main criteria for this selection were the stability of the system and the speed and accuracy of the control. The boundary conditions are derived from the UR5e manual and defined as follows:
\begin{equation*}
\begin{aligned}
&x_1 = \theta \in \begin{bmatrix} -6&6 \end{bmatrix};\\
&x_2 = \Dot{\theta} \in \begin{bmatrix} -3.14&3.14 \end{bmatrix};\\
&u \in \begin{bmatrix} -4.0&4.0 \end{bmatrix}
\end{aligned}
\end{equation*}

\subsection{Network topology}
Before setting up the simulation environment and the real test environment, the focus is on ensuring the remote control capabilities of the robot arm. For our network structure, a client-server model suits the requirements perfectly. The client-server system can be defined as a software architecture consisting of both a client and a server, where the clients always send requests while the server responds to the sent requests \cite{kratky2013client}. In the context of our NMPC system, the MPC controller plays the role of the server, while the robot arm is the client. As the client, the robot arm continuously sends its current state data to the server over the network at a specified frequency. The server, an MPC controller implemented on a remote computer, generates and sends control commands based on the information received from the client.

The network delay can be measured using RTT. However, since delay can occur in both forward and backward channels, the RTT delay is usually a sum of the delays in both channels, i.e. FWD and BWD. Since these two channels share the same physical cables and network links within the same communication network, it is difficult to really separate the two channels. However, considering that both channels share the same network communication configuration, a possible solution could be to change the queuing discipline of a switch to generate specific traffic such as delay and packet loss.

Local Area Networks (LANs) are usually built with network devices called switches that forward Ethernet frames between end hosts on the network. However, an alternative to using a physical switch is to use a Linux Ethernet Bridge. This software-based solution, created using the GNU/Linux operating system, behaves like a virtual switch or bridge. This approach offers greater flexibility and cost efficiency compared to traditional hardware-based switches \cite{varis2012anatomy}.

    \begin{figure}[htb]
    \centering
    \includegraphics[width=1\linewidth]{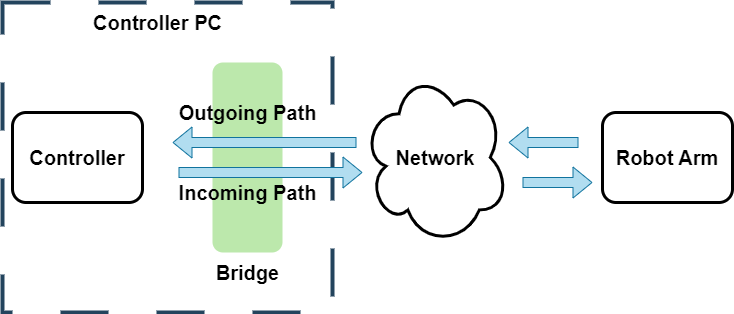}
    \caption{Linux Ethernet Bridge}
    \label{fig:chapter04:bridge}
    \end{figure}

As shown in Fig. \ref{fig:chapter04:bridge}, the Linux Ethernet Bridge is characterized by two distinct data paths: the outgoing path and the incoming path. Each path plays a specific role in managing traffic within the system. By changing the queuing discipline (or scheduling) of both the outgoing and incoming paths in an Ethernet interface, for example, by adding specific delay parameters or packet loss rates, we can manipulate the traffic generated on each Ethernet interface and path separately. 

\subsection{Buffering strategy}
The uncertainties of the network, such as network time delay and packet loss, can occur randomly at any point in the control process. The core principle of the buffer strategy is therefore to set up two buffers: one before the actuator and one before the controller. These buffers have both storage and comparison functionalities. When a new packet arrives, it is compared with the packets already stored in the buffer based on their timestamps. Only the packet with the most recent timestamp is used and stored in the buffer; all older packets are discarded. This ensures that the system always operates on the latest data and mitigates the effects of time delay and packet loss in the network.

\section{Results}
\label{Section3}
In this section, the performance of the implemented NMPC system is evaluated. First, an investigation to determine a suitable prediction horizon is presented. In order to analyze the robustness of the system under different network uncertainties such as network time delay and packet loss, several test methods were used. These tests aim to illuminate various aspects of the system's behavior and performance to provide a comprehensive evaluation of its effectiveness.

\subsection{Prediction Horizon Investigation}
Determining an appropriate prediction horizon for MPC is a complex task in the controller design phase. We define a suitable evaluation criterion to effectively evaluate the performance of the system. Integral Square Error (ISE) is used for this purpose:

\begin{equation*}
\begin{aligned}
ISE &= \int\varepsilon^2  dt 
= \sum_{i=1}^{n} \varepsilon^2 \\
\varepsilon &= x_{1}-x_{1,target}
\\
\Rightarrow ISE &= \frac{1}{6 \cdot n} \cdot \sum_{j=1}^{6} \sum_{i=1}^{n} (x_{j,1}(i)-x_{j,1,target}(i))^2
\end{aligned}
\end{equation*}
where $n$ denotes the total number of samples, $j$ denotes the number of joints, while $x_{j,1}$ and $x_{j,1,target}$ represent the current and target positions of the $j$-th joint, respectively. $\varepsilon$ signifies the error, which is the deviation of the current joint angle from the target angle. The ISE is calculated by integrating the square absolute joint state error over all six joints and over the entire sampling time. In this study, a low ISE value represents a controller with a lower deviation from the target, which means better system performance and higher precision and consistency.

\begin{figure}[htb]
\centering
\includegraphics[width=0.8\linewidth]{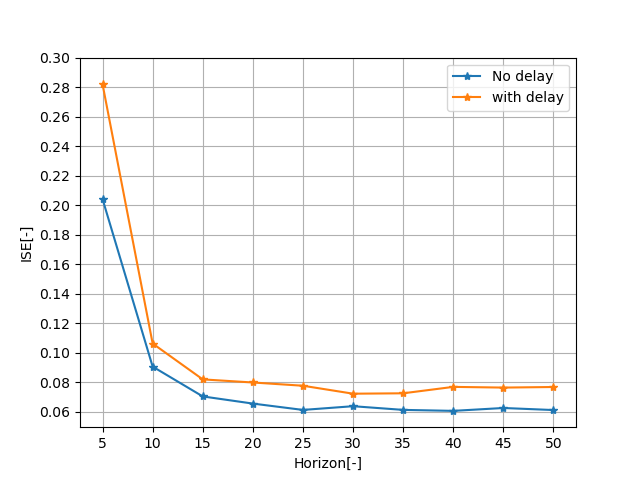}
\caption{Performance analysis over various prediction horizons}
\label{fig:horizon1}
\end{figure}

We apply various prediction horizons to the controller to include the effects of time delay in the network on both the forward and backward channels $100ms$ each. Observing the ISE value in Fig.~\ref{fig:horizon1} shows a decreasing pattern at the beginning, which stabilizes after increasing the horizon. The ISE value at higher horizons indicates that increasing the prediction horizon can not lead to significant improvement in the system performance. To select the optimal prediction horizon for the NMPC system, a more thorough analysis was performed for horizons 30 and 35. Our observations indicated that a horizon of 30 optimizes the performance and resilience of the NMPC system under conditions of network uncertainty. 

\subsection{Network Time Delay Analysis}
The performance of the NMPC system considering the effects of network time delay is analyzed here using both simulations and tests with real data. To analyze the robustness of the system, test methods such as multi-step response and sine wave response were used.

\subsubsection{Multi-step Response}
The multi-step response represents the system's reaction to a sequence of control commands over a period of time, which is useful for evaluating the stability, adaptability, and overall performance of the system in the face of varying inputs. The NMPC system is analyzed here under a total RTT delay of $200ms$, composed equally of $100ms$ forward delay and $100ms$ backward delay.

\begin{figure}[htb]
\centering
\includegraphics[width=0.8\linewidth]{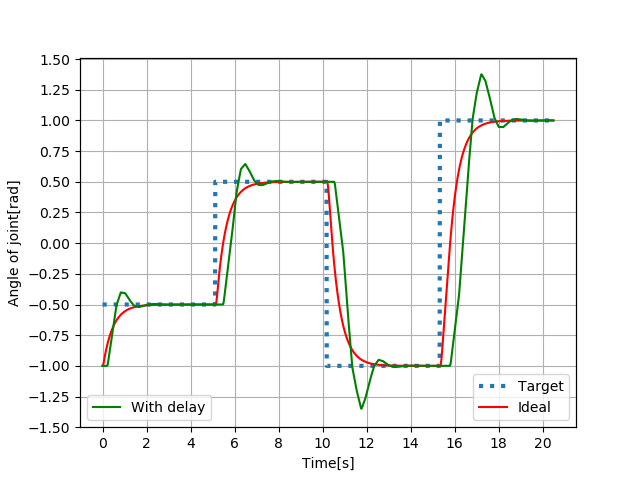}
\caption{Multi-step response of the NPCS}
\label{fig:multi}
\end{figure}

As depicted in Fig.~\ref{fig:multi}, both the magnitude of the overshoot and the delay of the control response both increase as the difference between the starting angle and the target angle increases. This indicates that the response of the system is strongly influenced by the discrepancy between the starting point and the desired endpoint. The larger this discrepancy, the more challenging it is for the system to reach the target quickly and accurately, especially in the presence of network delay.

\subsubsection{Sine Wave Motion}
\begin{figure}[htb]
\centering
\includegraphics[width=0.8\linewidth]{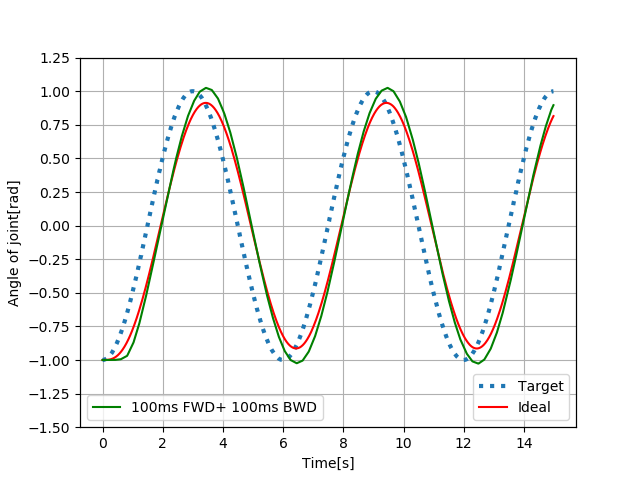}
\caption{Sine wave motion}
\label{fig:sinwave}
\end{figure}

A critical aspect of the system's performance evaluation is the operation of the robot arm during continuous motions such as a sine wave motion. Fig.~\ref{fig:sinwave} illustrates the effect of network time delay, including $100ms$ forward delay and $100ms$ backward delay, on the performance of the NMPC system during the sine wave motion of a single joint. This shows similar behavior to the multi-step response. In addition, a persistent phase shift is observed indicating a delayed response of the control output due to the network delay.

In order to make a meaningful comparison of the response of the sine wave motion to the effects of time delay in the network, it is important to determine appropriate performance metrics for robustness and efficiency. Different controllers have unique characteristics due to their respective configurations. Similar to the ISE, the evaluation metric known as the Residual Sum of Squares (RSS) can be used:

\begin{equation*}
 RSS=\frac{1}{n} \cdot \sum _{i=0}^{n-1}(x_1(i)-x_{1,ideal}(i))^{2}
\end{equation*}

In this study, RSS is used as an evaluation measure to quantify the difference between the actual and ideal state of the system. Here, $n$ denotes the total number of samples, while $x_1$ and $x_{1,ideal}$ represent the actual and ideal states of the system, respectively. The closer the value of RSS is to zero, the closer the actual state of the system is to the ideal state, which means less influenced by network uncertainties.

\begin{figure*}
\subfigure[simulation]{
\includegraphics[width=0.33\linewidth]{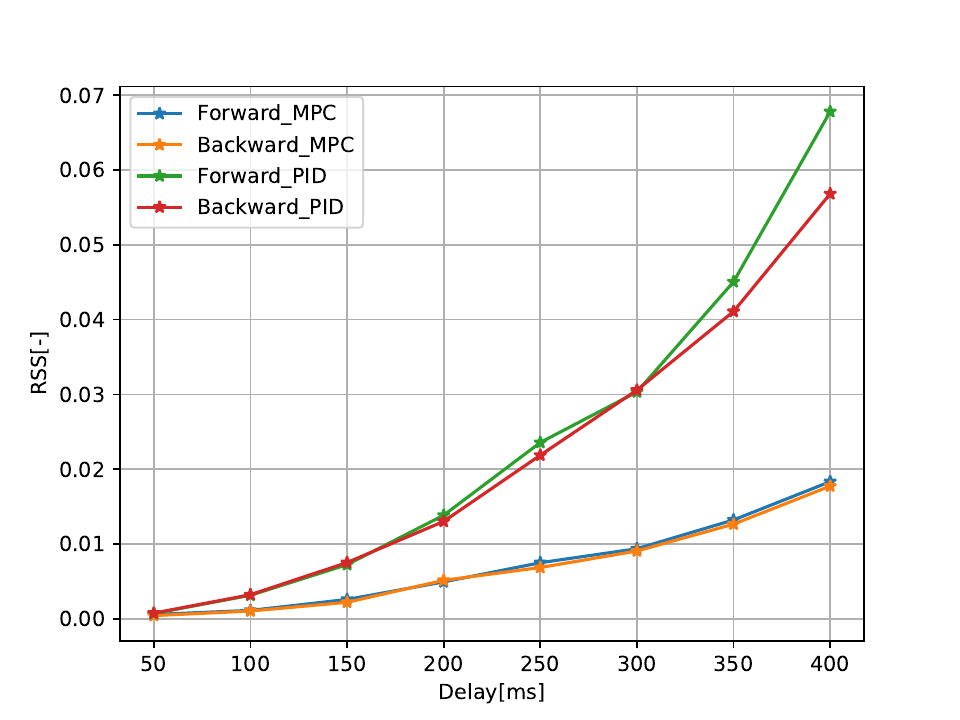}
\label{fig:sinwave_sim}
}
\subfigure[real robot]{
\includegraphics[width=0.33\linewidth]{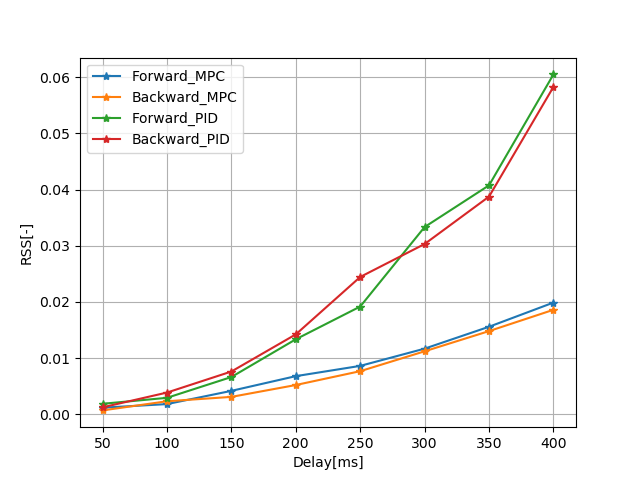}
\label{fig:sinwave_real}
}
\subfigure[real traffic]{
\includegraphics[width=0.33\linewidth]{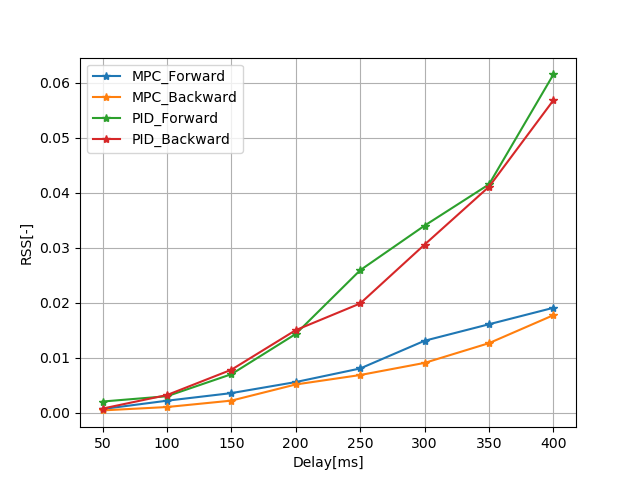}
\label{fig:realdata}
}
\caption{Performance comparison of NMPC system and NCS using PID in forward and backward channels}
\label{fig:MPCPID}
\end{figure*}

\paragraph{Comparison of NMPC system and an NCS using PID}
In order to thoroughly evaluate the robustness of the NMPC system in the face of network uncertainties, a comparative analysis with an NCS using a PID controller was performed. Such a comprehensive comparison, as shown in Fig.~\ref{fig:sinwave_sim}, reveals the particular characteristics and potential advantages of our system over another controlling system.

Moreover, validation tests with a real robot arm, as shown in Fig.~\ref{fig:sinwave_real}, have the same results as the simulation tests. Compared to the simulation results, the same sensitivity of our NMPC system to the forward delay and better robustness of NMPC to NCS using PID can be observed even in the presence of network delay. 

To test the NMPC system under real network traffic conditions, a traffic generator is used on both the controller side and the plant side, which continuously sends background packets to the other side. Due to the limited network bandwidth, the priority packets reach their destination with some delays. An evaluation of the performance of the NMPC system under these conditions is shown in Fig.~\ref{fig:realdata}. It should be noted that the time delay was measured using RTT, since the traffic in one channel may also affect the other. Therefore, the actual forward and backward delay will be slightly lower in this test. The results here are consistent with previous results from simulation tests and real robot tests.

\begin{figure}[htb]
\centering
\includegraphics[width=0.85\linewidth]{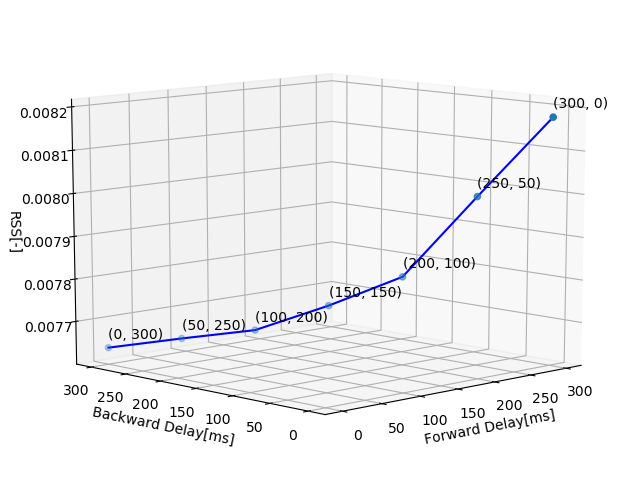}
\caption{Impact of varied delay distribution on system performance}
\label{fig:combi}
\end{figure}

\paragraph{Comparison of forward and backward delay}
To further investigate the different influences of forward and backward delays, another evaluation of the NMPC system was performed under scenarios with different combinations of these delays, as shown in Fig.~\ref{fig:combi}. In this evaluation, a constant total RTT delay of $300ms$ is maintained, but this delay is distributed differently between the forward and backward channels. This approach aims to more accurately determine how the NMPC system responds to different distributions of network delay, and thereby gain a deeper understanding of the separate effects of forward and backward delays on system performance.

A clear trend can be seen where the RSS curve reaches its lowest point at a backward delay of $300ms$ and peaks at a forward delay of $300ms$. It can be observed that the RSS value increases as the proportion of forward delay to the total delay increases. This demonstrates that the robustness and performance of the system is more affected by forward delays. This result supports previous evaluations that emphasize that the system is more sensitive to forward delays. Furthermore, the observations highlight the robustness and adaptability of the NMPC system, even under challenging network conditions with high delays.

\subsection{Packet Loss Analysis}
\begin{figure}[htb]
\centering
\includegraphics[width=0.85\linewidth]{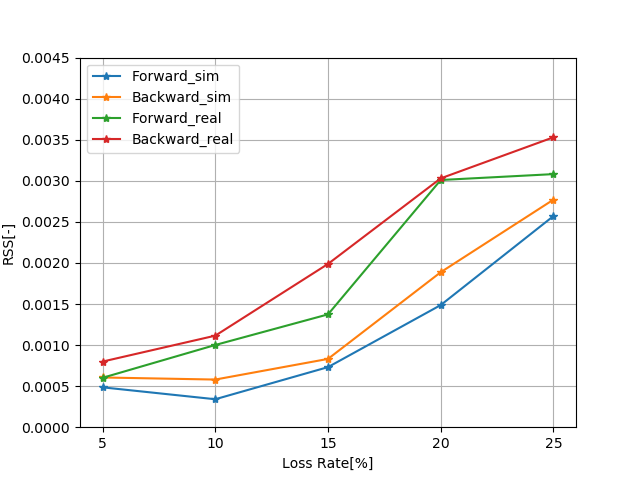}
\caption{Performance of sine wave motion with packet loss}
\label{fig:lossRSS}
\end{figure}

To gain a more comprehensive understanding of the performance of our NMPC system in the presence of packet loss, sine wave motion tests are performed in both simulated and real environments in Fig.~\ref{fig:lossRSS}.

The RSS values in this figure increase with increasing packet loss rates for both the forward and backward channels. This means that as the packet loss rate increases, the deviation of the actual state of the system from the ideal state grows, indicating a degradation in performance. It can be seen that packet loss seems to affect the backward channel more than the forward channel. One possible explanation for this could be that the loss of updated data for the MPC controller results in a lack of new input for the robot arm. Without these new inputs, the ability of the controller to effectively control the motion of the robot arm is compromised. The real-world test results show the same trends as the simulation results, further supporting these findings and demonstrating their relevance to real-world scenarios.

\subsection{Network Time Delay and Packet Loss Interaction}
\begin{figure}[htb]
\centering
\includegraphics[width=0.85\linewidth]{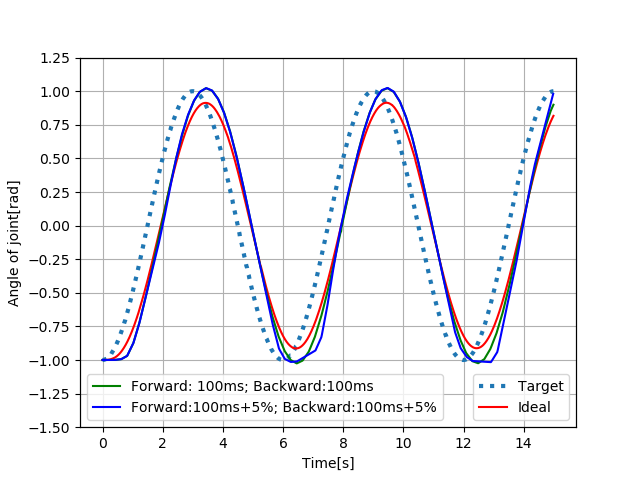}
\caption{Sine wave motion with network time delay and packet loss}
\label{fig:mix}
\end{figure}

In this study, the performance of the sine wave motion of the robot arm within the NMPC system is examined under network conditions that include both time delay and packet loss. In Fig.~\ref{fig:mix}, a comparison is shown between scenarios where only the network time delay is present and those where both the delay and packet loss are present:
\begin{itemize}
    \item  $100ms$ delay in the forward and backward channel without packet loss
    \item $100ms$ delay and $5\%$ packet loss rate in the forward and backward channel
\end{itemize}

It is worth noting that the control performance degrades with the introduction of network time delay. However, when packet loss is added, the control performance deteriorates further. This observation is particularly evident in the valleys of the sine wave. Compared to scenarios in which only a network delay occurs, a scenario in which a network time delay and packet loss coexist shows increased overshoot and a pronounced delay in control performance. This analysis indicates that the uncertainties of the network have a negative impact on the control performance.

\section{Conclusion}
\label{Section4}
In this paper, a networked model predictive control system has been implemented to stabilize the control of a robot arm. The performance of the system under various network conditions and uncertainties such as network time delay and packet loss is thoroughly analyzed. The structure of this system is first outlined with the inclusion of a buffering strategy to cope with network uncertainties. A model predictive controller is then designed based on a defined linear model of the robot arm and an understanding of the inherent characteristics and constraints of the robot arm. Considering the occurrence of delay in the forward and backward channels, a method of traffic generation was proposed for each channel separately, allowing a more detailed study of network influences. Extensive testing was performed, both in simulation environments and with real traffic, including a real robot arm. The method outlined in our implementation was found to be more robust to network influences compared to the traditional PID controller. These observations held for different delay conditions and were further substantiated by using the residual sum of squares as an evaluation measure.

\bibliographystyle{IEEEtran}
\bibliography{references}

\end{document}